\pdfoutput=1

\documentclass[11pt]{article}

\usepackage{EMNLP2023}
\usepackage{times}
\usepackage{latexsym}
\usepackage{booktabs}
\usepackage{xcolor}
\usepackage{multirow}
\usepackage[T1]{fontenc}

\usepackage[utf8]{inputenc}

\usepackage{microtype}

\usepackage{inconsolata}
\usepackage{graphicx}
\usepackage{algorithm}
\usepackage[noend]{algpseudocode}
\usepackage{amssymb}
\usepackage{xspace}
\usepackage{amsmath}
\usepackage[capitalize,noabbrev]{cleveref}
\usepackage{diagbox}
\usepackage{tabularx}
\usepackage{latexsym}
\usepackage{pifont}
\newcommand\tablecmark{\centering\color{green}\ding{51}}
\newcommand\tablexmark{\centering\color{red}\ding{55}}

\title{TOD-Flow: Modeling the Structure of Task-Oriented Dialogues}

\author{
Sungryull Sohn$^{1*}$ \hspace{0.5em} Yiwei Lyu$^{2*}$ \hspace{0.5em} Anthony Zhe Liu$^2$ \hspace{0.5em} Lajanugen Logeswaran$^1$ \\[0.2em]
{\bf Dong-Ki Kim}$^1$ \hspace{0.5em} {\bf Dongsub Shim}$^1$ \hspace{0.5em} {\bf Honglak Lee}$^{1,2}$ \\[0.2em]
$^1$LG AI Research \hspace{1em} $^2$University of Michigan, Ann Arbor
}

\newcommand{\todo}[1]{}
\renewcommand\todo[1]{\textcolor{red}{#1}}

\def\sact{a^r_t}

\def\ccond{\texttt{Can}\xspace}
\def\scond{\texttt{Shd}\xspace}
\def\sncond{\texttt{Shdnt}\xspace}
\def\cscond{\ccond\wedge\neg\sncond\xspace}

\newcommand{\subtask}[1]{\texttt{#1}}

\def\pfuncarg#1{f_{#1}}

\def\cfunc{f^{\ccond}}
\def\sfunc{f^{\scond}}
\def\snfunc{f^{\sncond}}
\def\csfunc{f^{\cscond}}
\def\cobj{J_{\ccond}}
\def\sobj{J_{\scond}}
\def\snobj{J_{\sncond}}
\def\csobj{J_{\cscond}}
\def\clab{y^{\ccond}_n}
\def\slab{y^{\scond}_n}
\def\snlab{y^{\sncond}_n}
\def\olab{a[n]}
\def\comp{\mathbf{c}}
\def\opt{\mathbf{a}}
\def\dialdata{\mathcal{D}}
\def\gdata{\mathcal{D}_{G}}
\def\nth{$n^\text{th}$\xspace}

\def\bcfunc{f^\text{BC}}

\def\woz{MultiWOZ\xspace}
\def\sgd{SGD\xspace}

\def\flan{FLAN-T5\xspace}
\def\gpt{GPT-turbo\xspace}

\def\ours{\tb{TOD-Flow}\xspace}
\def\ourstext{TOD-Flow\xspace}
\def\bcilp{\tb{BC}\xspace}
\def\ccilp{\tb{MSG$^2$}\xspace}

\def\galaxy{GALAXY}
\def\galaxys{GALAXY$^{*}$\xspace}
\def\hdno{HDNO}
\def\hdsa{HDSA}

\def\bleu{\textit{BLEU}\xspace}
\def\success{\textit{Succ}\xspace}
\def\inform{\textit{Info}\xspace}
\def\score{\textit{Score}\xspace}

\newcommand{\mb}{\mathbf}
\newcommand{\tb}{\textbf}
\newcommand{\mbb}{\mathbb}
\newcommand{\mc}{\mathcal}

\DeclareRobustCommand\onedot{\futurelet\@let@token\@onedot}
\def\onedot{.}
\def\eg{\emph{e.g}\onedot} 
\def\ie{\emph{i.e}\onedot}

\begin{document}
\maketitle
\def\thefootnote{*}\footnotetext{Equal Contribution}\def\thefootnote{\arabic{footnote}}
\begin{abstract}
Task-Oriented Dialogue (TOD) systems have become crucial components in interactive artificial intelligence applications. While recent advances have capitalized on pre-trained language models (PLMs), they exhibit limitations regarding transparency and controllability. To address these challenges, we propose a novel approach focusing on inferring the \ourstext graph from dialogue data annotated with dialog acts, uncovering the underlying task structure in the form of a graph. The inferred \ourstext graph can be easily integrated with any dialogue model to improve its prediction performance, transparency, and controllability. Our \ourstext graph learns what a model can, should, and should not predict, effectively reducing the search space and providing a rationale for the model's prediction.
We show that the proposed \ourstext graph better resembles human-annotated graphs compared to prior approaches.
Furthermore, when combined with several dialogue policies and end-to-end dialogue models, we demonstrate that our approach significantly improves dialog act classification and end-to-end response generation performance in the MultiWOZ and SGD benchmarks. Code available at: \url{https://github.com/srsohn/TOD-Flow}
\end{abstract}
\section{Introduction}\label{sec:intro}

\begin{figure*}[t]
    \centering
    \vspace{-0.5em}
    \includegraphics[clip, trim=0.0cm 10.7cm 11.7cm 0.0cm, width=1.02\textwidth]{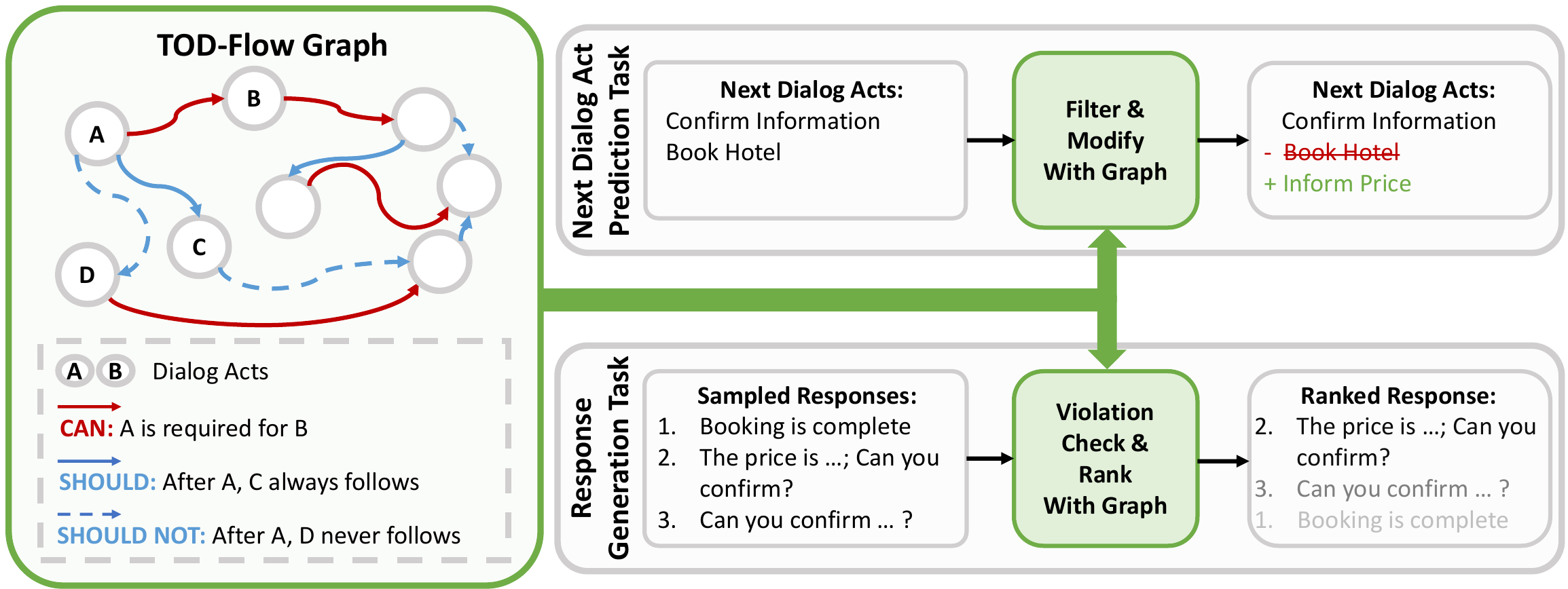}
    \caption{(Left) Our \ourstext graph captures the causal dependency between dialog acts in terms of {\color{red}can}, {\color{blue}should}, and {\color{blue} should not} relationships. 
    (Right)
    Intuitively, given the \ourstext graph, we can predict the relevant and irrelevant dialog act or responses based on the current dialog state. Based on the predicted relevance, we can filter and rank the base model outputs to enhance the prediction performance in both dialog act classification and end-to-end response generation tasks.}
    \label{fig:intro}
\end{figure*}

Task-Oriented Dialogue (TOD) systems have attracted significant attention due to their potential applications in personal assistants, customer support, and other interactive systems that necessitate human-like conversation~\cite{balaraman-etal-2021-recent,zhang2020recent}. 
Many of the recent advances in TOD have heavily leaned on pre-trained language models (PLMs)~\cite{he2022galaxy,wu2020tod} that are first pre-trained on a large corpus of data in an unsupervised manner, and then either fine-tuned~\cite{he2022galaxy, chen2019semantically, wang2020modelling} or subjected to few-shot prompting~\cite{hudevcek2023llms,labruna2023unraveling} to adapt them to specific dialogue domains. 
While these approaches have yielded commendable performance, they have limitations.
Few-shot prompted models have been challenged by issues of transparency, controllability, and adaptability to specific domains, especially when working with only a few examples.
The lack of understanding of their decision-making processes and fine-grained control over their output is often inadequate, which can result in sub-optimal conversational experiences.
On the other hand, fine-tuned models are confronted with their own unique challenges. While they offer improved performance by aligning the model with task-specific semantics, this approach typically requires large annotated datasets and resources which can be a limiting factor in practice. Furthermore, these models often lack transparency, making it challenging to understand the reasons behind their decisions or predictions.

Some prior works~\cite{raghu2021end,laradji2023workflow} introduced workflow-based dialog models to handle the challenges in existing TOD models.
These methods aim to explicitly model the structure of dialog in a graph format.
Grounding the dialog in the graph offers benefits in terms of 1)
elucidating the reasoning of system's decisions in terms of the relationships (\ie, transparency); 2) allow human manipulation of the dialogue model via graph modification (\ie, controllability) without retraining the dialog model.
However, real-world dialogues were often unstructured, making it non-trivial to be modeled as a workflow, and the necessity of manually designing the domain-specific workflow or its elements limits its practical applicability.

To tackle these challenges, we propose to learn the \emph{subtask graph}~\cite{sohn-neurips18} from task-oriented dialog data.
Intuitively, the subtask graph can predict the affordance (\ie, availability) of the action from the status of environment and agent (\ie, the progress of completing a task or the subtasks). 
The subtask graph framework has two major benefits: i) subtask graphs can be inferred from the demonstrations without any direct supervision (\eg, video~\cite{jang2023multimodal} or transcript~\cite{logeswaran2023unsupervised}), 
ii) subtask graphs can be combined with the base prediction model to improve its prediction since the subtask graph does not decide what to predict but instead suggests the affordable candidates of prediction.

\paragraph{Contributions.}
The main contribution of this work is generalizing the subtask graph framework into task-oriented dialog settings.
To this end, we propose the \emph{\ourstext graph}, which extends the subtask graph framework in three major aspects.
First, we show that subtask graph can infer the relationship between dialog state and dialog acts without requiring any manual definition of nodes and edges in graphs.
Second, in addition to the precondition (or \emph{can} relationship), we present learning algorithms to model two novel relationships, \emph{should} and \emph{should not}, which provide more fine-grained control and improved prediction.
For instance, a \emph{can} relationship may represent that the system can make a payment only if the user confirms the payment.
The \emph{should} relationship may learn that if a user ask about the address of the hotel, the system should reply back.
Conversely, a \emph{should not} relationship may dictate that the system usually does not predict a farewell if the user's last utterance implies a question.
Third, we demonstrate that the inferred \ourstext graphs can enhance any dialog policy or end-to-end dialog system, whether fine-tuned or prompted, without the necessity of retraining.
\section{Background}\label{sec:background}
Our main contribution, the \ourstext graph, is an extension of the subtask graph framework~\citep{sohn-neurips18, sohn-iclr20}, which describes the causal dependency structure of a compositional task $\tau$ consisting of a set of subtasks.
In the context of task-oriented dialogue, dialog acts can be seen as subtasks.
Each subtask has a \textbf{precondition} that must be satisfied before the subtask can be performed.
Note that the precondition is not the only relationship between subtask, and in~\Cref{subsec: 32} we extend it by incorporating other types of relationships.
Since precondition describes the causal relationship between subtasks, it imposes a constraint on the order in which subtasks can be performed (\eg, the system can make a payment \emph{only after} the user confirms the payment).
Formally, we define the precondition as a Boolean expression consisting of Boolean constants (\eg, True or False), Boolean variables and logical connectives (\eg, and ($\&$), or (\( \mid \))). 
To illustrate, consider the precondition of subtask \subtask{C}: $\pfuncarg{\subtask{C}} = \&(\subtask{A}, \subtask{B})$, where the subtasks \subtask{A} and \subtask{B} must be completed before \subtask{C} is completed.
It can be equivalently viewed as a Boolean function where inputs are Boolean variables indicating whether subtasks \subtask{A} and \subtask{B} are completed, and the output represents whether the precondition $\pfuncarg{\subtask{C}}$ is satisfied: $\pfuncarg{\subtask{C}}(\subtask{A}=\text{True},\subtask{B}=\text{False}) = \text{True}~\&~\text{False} = \text{False}$.
Also, the boolean expression $\pfuncarg{\subtask{C}} = \&(\subtask{A}, \subtask{B})$ can be viewed as a \textbf{graph} with vertices consisting of subtasks and logical operators $V = \{\subtask{A}, \subtask{B}, \subtask{C}, \&\}$ and edges $E=\{\subtask{A} \rightarrow \&$, $\subtask{B} \rightarrow \&$, $\& \rightarrow \subtask{C}\}$ that represent preconditions.
We will use these different views of the precondition (\ie, as a boolean expression, graph or function) interchangeably.
The \textbf{subtask graph} visualizes the preconditions $\pfuncarg{1},\ldots$ of the subtasks (see \Cref{fig:intro} for examples).
We note that the subtask graph has been adopted in various settings~\cite{liu-aaai22, sohn-iclr20, sohn-uai22} and subsumes other task graph formats~\cite{andreas-icml17, boutilier-ijcai95, sakaguchi-emnlpf21}, flowchart~\cite{raghu2021end}, and workflow~\cite{laradji2023workflow}.
\section{\ourstext Graph Learning}

\subsection{Problem Formulation}
\label{sec:problem}
For dialogue turn $t$, let $u_t$ be the user input and $r_t$ be the corresponding system response.
The user inputs and system responses are represented as a set of \emph{dialog acts} that labels the raw language utterance at each turn according to its category of meaning: $u_t, r_t \subset \mc{A}$, where $\mc{A}$ is the set of dialog acts.
Let $d_t$ be the database query result that can be obtained through querying the database.
Then, the dialog data $\mc{D}$ is a set of dialog trajectories $\mc{D}^\tau=\{(u_0, r_0, d_0, u_1, \ldots), \ldots \}$. 
Given the dialog data $\mc{D}^\tau$, the goal is to generate the \emph{\ourstext graph} $G$ that models the dependency between system acts, user acts, and database results in graph format.
\paragraph{Challenges.}
There are two main challenges in tackling this task.
First, the information in the dialogue is noisy
due to annotation errors such as missing or ambiguous dialog acts, slots, and values.
Second, these dialog annotations only provide partial information about the underlying relationships between subtasks.
Thus, we need to infer whether each relationship is satisfied or not from the dialogue annotations.
We describe how we overcome these challenges in \Cref{subsec:graph_inference}.

%

\subsection{\ourstext Graph}
\label{subsec: 32}
For each dialog act $\opt$, the \ourstext graph is defined in terms of three conditions: $\ccond_\opt$, $\scond_\opt$, and $\sncond_\opt$.
Intuitively, $\ccond_\opt$, $\scond_\opt$ and $\sncond_\opt$ condition respectively defines whether the dialog act $\opt$ \emph{can}, \emph{should}, and \emph{should not} be performed by the agent (user or system) at a given status.
Similar to the precondition in subtask graph framework (see \Cref{sec:background}), each condition is defined as a Boolean expression.
Also, it can be equivalently viewed as a Boolean function $\cfunc, \sfunc, \snfunc: \comp \mapsto \{0, 1\}$ or a graph (see \Cref{sec:background}), where $\comp\in\{0, 1\}^{N_\tau}$ is the subtask completion (or dialog state) vector indicating whether \nth subtask has been achieved (\ie, $c[n]=1$) or not (\ie, $c[n]=0$).
%
%

\subsection{Learning \ourstext Graphs}
\label{subsec:graph_inference}
\paragraph{Dataset.}
Given the dialogue data $\mc{D}=\{(u_t, d_t, r_t)\}$, we aim to build the graph inference dataset $\gdata=\{(\comp_t, \opt_t)\}$, from which we can infer the \ourstext graph $\cfunc, \sfunc$, and $\snfunc$.
The action set $\opt_t$ is the set of dialog acts that were performed at turn $t$: $\opt_t = u_t \cup r_t$.
The completion set $\comp_t$ is the set of dialog acts and database query that has ever been performed before turn $t$: $\comp_t = \comp_{t-1}\cup \opt_{t-1} \cup \{d_{t-1}\}$.

\noindent
\begin{table}
\begin{tabular}{|c||*{2}{c|}}\hline
\diagbox{Cond}{$\olab$}
& \begin{tabular}{@{}c@{}}Executed \\ $(\olab=1)$\end{tabular}
& \begin{tabular}{@{}c@{}} Not executed \\ $(\olab=0)$\end{tabular}\\\hline\hline
$\sfunc_n=1$ & True positive & False positive \\\hline
$\sfunc_n=0$ & - & - \\\hline\hline
$\snfunc_n=1$ & False positive & True positive \\\hline
$\snfunc_n=0$ & - & - \\\hline\hline
$\cfunc_n=1$ & True positive & - \\\hline
$\cfunc_n=0$ & False negative & - \\\hline\hline
$\bcfunc_n=1$ & True positive & False positive \\\hline
$\bcfunc_n=0$ & False negative & True negative \\\hline
\end{tabular}
\caption{Confusion matrix of the \ourstext graphs (\scond, \sncond, \ccond) and the BC baseline with respect to the dialog act label $\olab$. The empty cell ($-$) indicates that the label for each relationship is unavailable.}
\label{tab:confusion}
\end{table}


\paragraph{\scond inference.}
When should condition is satisfied (\ie, $\sfunc_n(\comp)$=1), the agent is required to perform the \nth dialog act (\ie, $\olab=1$). When the should condition is not satisfied (\ie, $\sfunc_n(\comp)$=0), the should relationship has no effect on the policy. This relationship can be represented as the confusion matrix shown in the first two rows in~\Cref{tab:confusion}.
Accordingly, we maximize the true positive, while minimizing the false positive by maximizing the objective $\sobj$ in \Cref{eq:sobj}.
\begin{align}
    \sobj &=
    \mbb{E}_{(\comp, \olab)}
    \left[
        \mbb{I}(\olab = 1 |\sfunc_n(\mb{c})=1)
    \right]\label{eq:sobj}
\end{align}

\paragraph{\sncond inference.}
Similar to \scond, when the shoud-not condition is satisfied, (\ie, $\snfunc_n(\comp)=1$), the agent is required \emph{not} to perform the \nth dialog act (\ie, $\olab=0$), and agent has a freedom to execute the \nth dialog act when the condition is not satisfied (\ie, $\snfunc_n(\comp)=0$).
\Cref{tab:confusion} summarizes the relationship between $\sncond$ and $\olab$.
We learn $\snfunc_n$ by maximizing the following objective:
\begin{align}
    \snobj &= 
    \mbb{E}_{(\comp, \olab)}
    \left[
        \mbb{I}(\olab = 0 |\snfunc_n(\mb{c})=1)
    \right].\label{eq:snobj}
\end{align}

\begin{figure}[!t]
\centering
\includegraphics[width=0.45\textwidth]{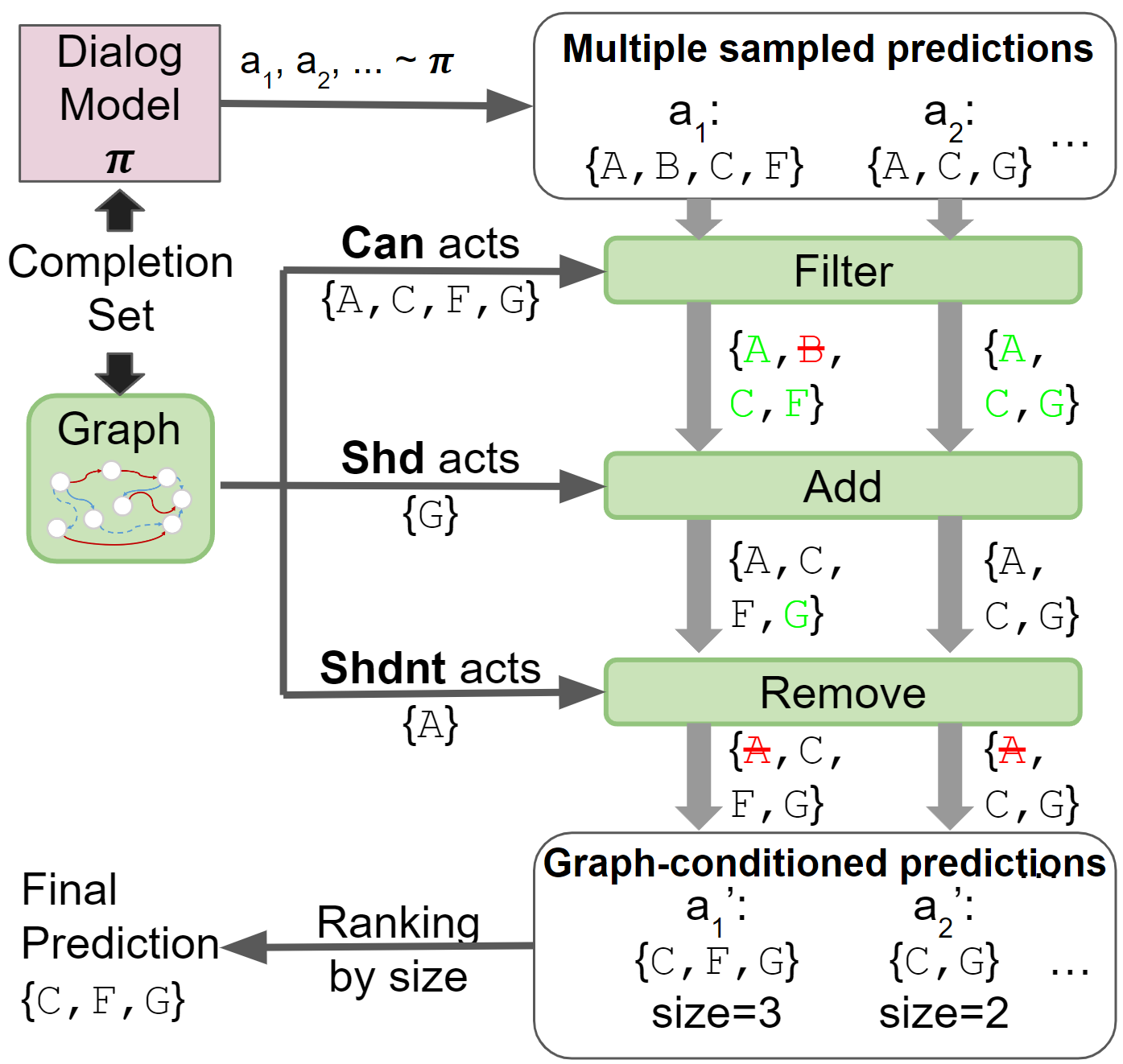}
\caption{An illustration of our graph-conditioned dialog policy. We sample multiple predictions from base dialog policy $\pi$, then use our graph to filter, add and remove acts from each prediction following \Cref{alg:graph_conditioned_DM}. Then we obtain final prediction by selecting the graph-conditioned prediction with the largest size.}
\label{fig:filter}
\end{figure}

\paragraph{\ccond inference.}
By definition of \ccond (or precondition), a dialog act can only be performed ($\opt_n=1$) if its \ccond condition is satisfied (\ie, $\cfunc_n(\comp)=1$): a true positive case in~\Cref{tab:confusion}. 
On the contrary, it is a contradiction if a dialog act $\opt_n$ is performed while its \ccond is not satisfied (\ie, $\cfunc_n(\comp)=0$): a false negative case in~\Cref{tab:confusion}.
Thus, the $\cfunc$ can be learned by maximizing the following objective:
\begin{align}
    \cobj &= \mbb{E}_{(\comp, \olab)}
    \left[
        \mbb{I} [\cfunc_{n}(\comp)=1 | \olab=1]
    \right].\label{eq:can-obj}
\end{align}
However, different from \scond and \sncond, inferring \ccond is nontrivial, because if we maximize $\cobj$, we get the trivial precondition: always true (\ie, $\cfunc_n(\comp) = 1$ for all $\comp$).
Previous works handled this issue by either making additional assumptions~\citep{hayes-icra16,huang2019neural} or applying regularization~\cite{jang2023multimodal}.
However, these approaches unavoidably introduces noise in learning, and require careful hyperparameter tuning to balance between objective and regularization.
Instead, inspired by the fact that \ccond and $\neg$\sncond (\ie, negation of \sncond) applies to the policy in the same manner 
(\ie, the dialog act $\opt$ can be performed if both $\cfunc_n=1$ and $f^{\neg\sncond}_n=1$),
we propose to infer \ccond and \sncond simultaneously as follows:
\begin{align}
    &\csobj \nonumber\\
    &= \mbb{E}_{(\comp, \olab)}
    \left[
        \mbb{I} [\csfunc_{n}(\comp)=1 | \olab=1] \right.\nonumber\\
    &\left. \quad+\ \alpha\ \mbb{I} [\csfunc_{n}(\comp)=0 | \olab=0]
    \right],\label{eq:csobj}
\end{align}
where $\alpha$ determines the relative weight between \ccond and \sncond in optimization.
Intuitively, the agent can perform \nth dialog act only if \ccond is satisfied and \sncond is not satisfied.

\paragraph{Baseline.} As an ablation model, we consider the behavioral cloning (BC)~\cite{michie1990cognitive} objective, which tries to mimic the demonstration behavior:
\begin{align}
    J_{\text{BC}} = \mbb{E}_{(\comp, \olab)}
    \left[
        \mbb{I}\left[\bcfunc_n (\comp)= \olab\right]
    \right]\label{eq:bcobj}
\end{align}
The confusion matrix is shown at the bottom of \Cref{tab:confusion}.

We can use any binary classification models to optimize the objectives (\ref{eq:sobj}), (\ref{eq:csobj}), and (\ref{eq:bcobj}). We used the decision tree models in the experiment following the previous works~\cite{boutilier-ijcai95, huang2019neural, sohn-iclr20}.

\section{Graph-conditioned Dialog Modeling}
\label{sec:gcond}
We describe how the inferred TOD-flow graph $G$ can enhance the prediction performance of any off-the-shelf dialogue policies and end-to-end dialog systems.

\begin{algorithm}[!t]
\caption{TOD-flow Graph-conditioned Dialogue Model}\label{alg:graph_conditioned_DM}
\begin{algorithmic}[1]
\Require Dialogue model $\pi$, TOD-flow graph $\cfunc, \sfunc, \snfunc$, Completion $\comp$
\Ensure Sampled dialog acts $\opt$
\State $\opt\sim\pi$ \Comment{Sample dialog acts from $\pi$}
\State $\opt \gets \opt \cup \{\opt'|\sfunc_{\opt'}(\comp)=1\}$ \Comment{Apply \scond}
\State $\opt \gets \opt \cap \{\opt'|\csfunc_{\opt'}(\comp)=1\}$ \\
\Comment{Apply $\cscond$}\\
\Return $\opt$
\end{algorithmic}
\end{algorithm}

\begin{figure*}[!t]
\centering
\includegraphics[width=0.95\textwidth]{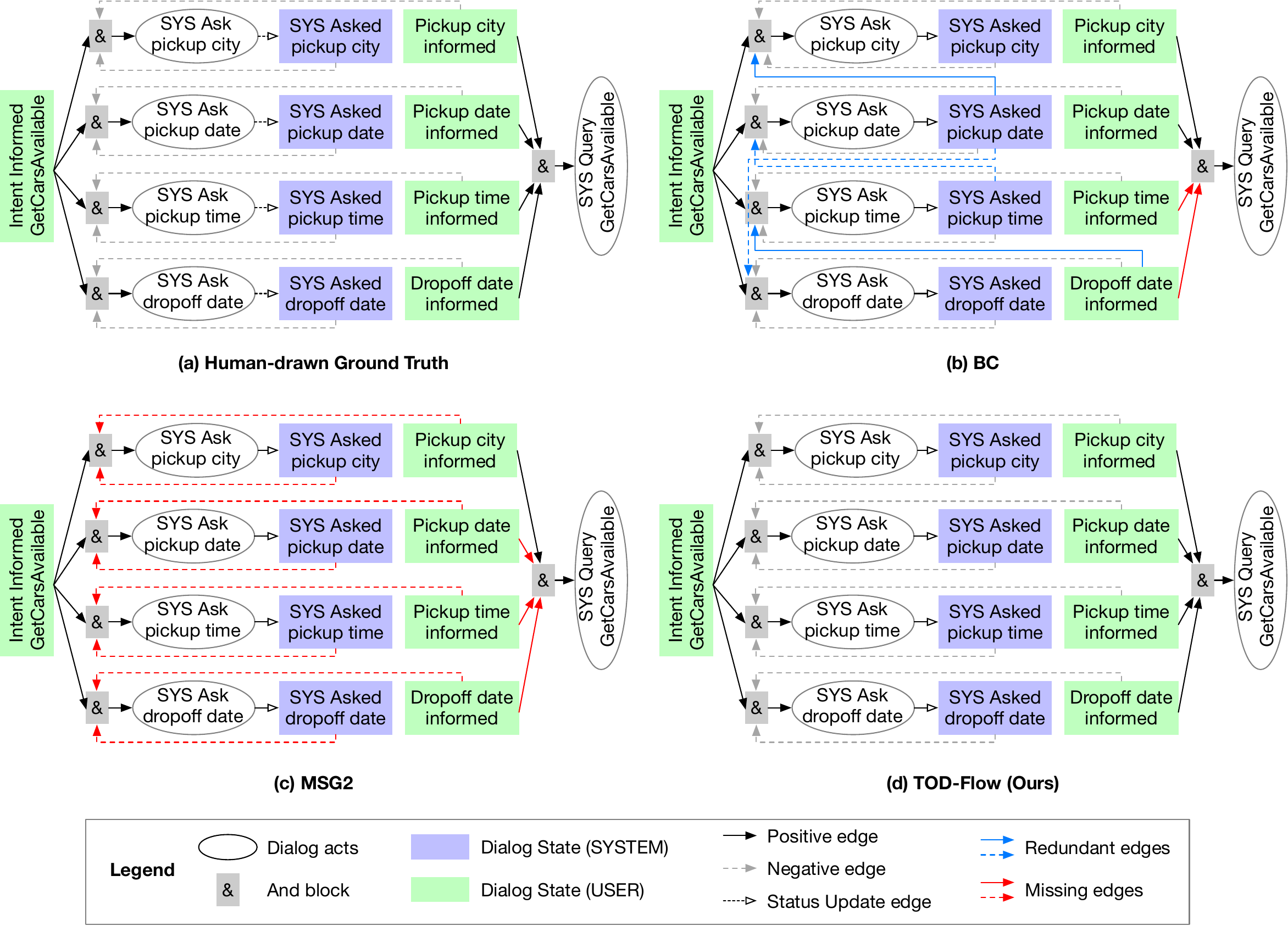}

\caption{Comparing baselines and our method against human-drawn ground truth graph on a subpart of the \texttt{RentalCars\_1} domain of SGD. Dotted lines are negative edges (logical negations). \textcolor{red}{Red edges} are missing edges and \textcolor{blue}{blue edges} are redundant edges compared to the ground truth. In this scenario, the system is allowed to ask for a slot if user has informed the intent of finding an available car and has not yet informed this slot and the system have not yet asked for the slot; and the system can only query for a car when all 4 slots are informed. Our method perfectly matches the ground truth, while both baselines have incorrect edges (\bcilp has 4 redundant edges and 2 missing edges, while \ccilp has 11 missing edges).}
\label{fig:qualitative}
\end{figure*}
\subsection{Graph-conditioned Dialog Policy}
\label{subsec:gcdp}

\vspace{-2mm}
The inferred TOD-flow graph (\scond and $\cscond$) can propose the dialog acts that can, should, and should not be performed given the current dialog history (or completion set).
\Cref{fig:filter} and \Cref{alg:graph_conditioned_DM} describes the entire process.
Given a base dialogue policy $\pi^\text{DP}$,
we sample the system acts from the policy $a\sim\pi^\text{DP}$, and filter, add and remove acts from the sampled system acts according to the \ccond, \scond, and \sncond, respectively.

We can further improve the prediction performance if our baseline dialog model can be sampled multiple times with different results, as illustrated in Figure~\ref{fig:filter}. We use the graph to condition each candidate result, then select the best one using a selection method such as most number of actions in set, candidate with least graph violations, etc. We empirically found that simply choosing the result with the most actions works best. 

\vspace{-2mm}
\subsection{Graph-conditioned End-to-end Response Generation}
\label{subsec:gce2e}

\vspace{-2mm}
End-to-end dialogue system directly reads and outputs the utterances in natural language form.
Given the base end-to-end model $\pi^\text{e2e}$, we sample multiple system utterances from the base model (mostly via beam-search alternates, see Appendix~\ref{app:alternates} for details).
Then, we use a few-shot prompted \texttt{GPT-3.5-turbo} model to annotate each generated candidate utterance with the dialog act (see Appendix~\ref{app:nlu} for details).
Finally, we use the inferred graph to choose the best utterance that has the least \emph{violation rate} (\ie, portion of dialog acts that violates the inferred \ccond, \scond, and \sncond conditions).
Note that with this approach, all final responses still comes from the base end-to-end model, so the improvement is still upper-bounded by the capabilities of the model. Our graph simply presents a better candidate selection method.

\begin{table*}[!t]
\centering
\scalebox{1.0}{
\begin{tabular}{lcccc}
\toprule
\multirow{2}{*}[-0.23em]{Models} & \multicolumn{2}{c}{\sgd (24 domains)} & \multicolumn{2}{c}{\woz (14 domains)}  \\
 \cmidrule(lr){2-3} \cmidrule(lr){4-5}
& FLAN-T5 & \gpt & FLAN-T5 & \gpt  \\ 
\midrule
No Graph& 49.9\%& 78.8\% &  21.6\%& 40.8\%\\
+\bcilp & 57.2\%& 71.8\% &  23.5\%& 38.2\%\\
+\ccilp~\citep{jang2023multimodal} & 52.4\%& 79.4\%  & 23.8\%& 40.2\%\\
+\ours (ours) & \textbf{83.1\%} & \textbf{89.2\%}  & \textbf{35.0\%} & \textbf{48.2\%} \\ 
\bottomrule 
\end{tabular}}
\caption{Average F-1 scores of next system action prediction experiment on two datasets (SGD and MultiWOZ) and two large language models (FLAN-T5 and \gpt) as few-shot predictors. We can see that while \bcilp often damages performance, \ours consistently improves performance by a significant amount.}
\label{tab:dm}
\end{table*}

\section{Experiments}\label{sec:experiment}

We perform experiments 
to show that
(1) the TOD-flow graph can be accurately predicted without any supervision,
(2) our graph can improve the accuracy of dialog policy models, 
and (3) our graph can improve the quality of response generation in end-to-end dialog models.

\subsection{Dataset}
We used two standard TOD benchmarks.
Schema-Guided Dialogue (\sgd)~\cite{rastogi2020towards} has over 20k task-oriented simulated conversations based on human-designed schema.
\sgd covers a wide range of domains (\ie, different dialog acts and goals).
We use 24 domains in SGD, and did not use the schema for experiment.
\woz~\cite{budzianowski2020multiwoz} has 10k human-human conversations on 14 domains.
Since \woz is collected from human-human conversations, the utterances are much more linguistically diverse than SGD.
Also, different from SGD where the annotations are generated from the schema, annotations in \woz are labeled by human.
Therefore, the annotations in \woz are often noisy (\ie, inconsistent, wrong, or missing), which present additional challenge compared to SGD.

For both datasets, we obtain train/test splits of the dialogs within each domain (see Appendix~\ref{app:dp} for details).
The training set is used for 1) inferring \ours graph, 2) building demonstration for few-shot prompted models, and 3) finetuning the finetuning-based models.
The test set is only used for evaluation.
For graph inference, we map the dialog act of user, database, and system to completion $\comp$ and dialog act $\opt$ vectors as described in~\Cref{subsec:graph_inference}.

\subsection{Baselines}
We compare three graph inference algorithms:
\begin{itemize}
    \item \bcilp learns to imitate the demonstration via behavioral cloning (see \Cref{subsec:graph_inference})
    \item \ccilp~\cite{jang2023multimodal} learns the subtask graph by optimizing the $\cobj$ (see~\Cref{eq:can-obj}) with complexity regularization.
    \item \ours (ours) is our \ours graph learning algorithm.
\end{itemize}
For fair comparison, we used the scikit-learn decision tree model~\citep{scikit-learn} for all the graph inference algorithms.

\subsection{TOD-Flow Graph Inference}

We first qualitatively compare the inferred graphs with the human-drawn graphs on \texttt{RentalCars\_1} domain in SGD dataset.
We found that in general \ours produces graphs that agree with the human-drawn graphs much more often compared to baselines (\bcilp and \ccilp). 
\Cref{fig:qualitative} illustrates the subpart of the inferred and human-drawn graph, where \ours inferred the graph perfectly matches the human-drawn graph, while the baselines missed important information (such as not requiring all 4 required slots to be informed before performing the query).

\begin{table*}[t]
\centering
\begin{tabular}{rccccccc}
\toprule
&\multirow{2}{*}[-0.23em]{\begin{tabular}[c]{@{}c@{}}\ccond-\sncond \\ graph\end{tabular}}  & \multirow{2}{*}[-0.23em]{\begin{tabular}[c]{@{}c@{}}\scond \\ graph\end{tabular}}& \multirow{2}{*}[-0.23em]{\begin{tabular}[c]{@{}c@{}} Ranking \\ method \end{tabular}}& \multicolumn{2}{c}{\sgd (24 domains)} & \multicolumn{2}{c}{\woz (14 domains)} \\
\cmidrule(lr){7-8} \cmidrule(lr){5-6}
 &  & & & FLAN-T5 & \gpt  & FLAN-T5 & \gpt \\ \midrule
No graph & \tablexmark & \tablexmark & Greedy & 49.9\%& 78.8\%&  21.6\%& 40.8\% \\
 \midrule
Graph& \tablexmark & \tablecmark & Compliance & 70.4\% & 82.5\% & 26.1\% & 41.6\%\\
ablations & \tablecmark & \tablexmark & Compliance&  65.6\% & 83.5\% & 30.4\% & 44.5\%\\ \midrule
\multirow{4}{*}{\begin{tabular}[l]{@{}c@{}}Sampling  \\ \& ranking \\ ablations\end{tabular}}  & \tablecmark & \tablecmark &Greedy &  80.9\% & 88.5\% & 26.2\% & 45.5\%\\
&\tablecmark &\tablecmark & Majority &  75.2\% & 88.1\% & 20.7\% & 45.0\%\\
&\tablecmark &\tablecmark & Violation &  76.2\% & 88.3\% & 23.7\% & 44.7\% \\
&\tablecmark &\tablecmark & Uniform &  75.8\% & 88.0\% & 23.2\% & 45.2\% \\
\midrule
Ours & \tablecmark &\tablecmark & Compliance & \textbf{83.1\%} & \textbf{89.2\%} & \textbf{35.0\%} & \textbf{48.2\%} \\

\bottomrule
\end{tabular}
\caption{Ablation studies on graphs and ranking methods for next system action prediction. The numbers shown are average F-1 scores. We see that both \ccond-\sncond and \scond graphs have significant contributions towards the performance and our Compliance ranking method outperforms greedy sampling or other ranking methods.}
\label{tab:dmablation}

\end{table*}

\subsection{Task 1: Dialog Policy Learning}


\paragraph{Base Models.}
We use two instruction-tuned large language models (LLM) as baseline dialog policy: \flan-xxl~\cite{chung2022scaling} and \gpt\footnote{https://platform.openai.com/docs/models/gpt-3-5}. 
At each turn, we prompt the LLM with five demonstration dialogues from train split of the same domain followed by the dialogue history, and ask the model to predict next system dialog acts. See Appendix~\ref{app:prompt} for more details on prompting the LLM.

\paragraph{Evaluation Protocol.}
We sample 10 candidate predictions from the base models, which is filtered and ranked based on the graph (see \Cref{subsec:gcdp}) to choose the best prediction. 
\paragraph{Metric.}
We measure F-1 score between the ground-truth and predicted system dialog acts at each turn and average over entire domains.

\paragraph{Results.}
Table~\ref{tab:dm} summarizes the F-1 score of each model on \sgd and \woz.
Overall, we observe that \ours consistently improves the prediction accuracy with a significant margin compared to other baselines \bcilp and \ccilp on all base models and all dataset.
We also found that the improvements are bigger on \flan compared to \gpt. 
This indicates that the \gpt already models the \ccond, \scond, and \sncond to some extent, so that augmenting it with the graph provide less benefits.
Since \bcilp learns to mimic the exact behavior in demonstration, \bcilp tends to dictate the base policy more aggressively and hurts the performance when combined with strong base model \gpt.
\ccilp correctly models the precondition of dialog acts, but provides less benefit compared to \ours due to the conservative graph learning (\ie, complexity regularization) and lacking the ability to model \scond and \sncond relations.

\paragraph{Ablations.}
To further justify our design choices, we performed ablation studies on two key components of \ours: the graphs and the ranking method after graph-conditioning. 
For filtering graphs, we examined the effect of \ccond-\sncond and \scond graphs. 
Regarding the ranking method, we compare the proposed ranking approach (\ie, Compliance) against various alternatives:
\begin{itemize}
    \setlength\itemsep{0.1em}
    \vspace{-1mm}
    \item \tb{Greedy} ranks by likelihood of base LLM.
    \vspace{-1mm}
    \item \tb{Compliance} ranks predictions by larger number of actions complying with the graph (\ie~rank by size in~\Cref{fig:filter}).
    \vspace{-1mm}
    \item \tb{Majority} chooses the majority prediction among the multi-sampled predictions.
    \vspace{-1mm}
    \item \tb{Violation} ranks predictions by least number of actions filtered, added, and removed in graph conditioning.
    \item \tb{Uniform} randomly chooses one of the multi-sampled predictions.
    \vspace{-1mm}
\end{itemize}
The results are shown in Table~\ref{tab:dmablation}, and \ours outperformed all ablations, showing the necessity of all graphs and ranking by largest set.

\begin{table}[!ht]
\setlength{\tabcolsep}{4.2pt}
\begin{tabular}{lccc|c}
\toprule
Graph& \bleu & \inform & \success & \score \\ \midrule
\multicolumn{5}{c}{HDSA} \\ \midrule
(no graph) & 20.74 & 87.20 & 78.00 & 103.34 \\
+\bcilp & \textbf{20.76} & 86.80 & 77.70 & 103.01 \\
+\ccilp & 20.71 & 86.90 & 77.90 & 103.11 \\
+$\text{\ours}^\dagger$ & 20.69 & 87.70  & 78.20 & 103.64 \\ 
+\ours & 20.51 & \textbf{88.10}  & \textbf{79.00} & \textbf{104.06} \\ \midrule
\multicolumn{5}{c}{HDNO} \\ \midrule
(no graph) & 17.83 & 93.00 & 84.50 & 106.58 \\
+\bcilp & 17.79 & 92.90 & 84.50 & 106.49 \\
+\ccilp & 17.83 & 92.90 & 84.40 & 106.48 \\
+$\text{\ours}^\dagger$ & \textbf{18.08} & 93.10 & \textbf{85.10} & \textbf{107.18} \\
+\ours & 17.97 & \textbf{93.20} & 85.00 & 107.07\\ \midrule
\multicolumn{5}{c}{\galaxy} \\ \midrule
(no graph) & \textbf{19.92} & 92.00 & 82.80 & 107.32 \\
+\bcilp & 19.85 & 91.80 & 82.50 & 107.00 \\
+\ccilp & 19.69 & 91.30 & 81.00 & 105.84 \\
+$\text{\ours}^\dagger$ & 19.86 & \textbf{92.40} & 83.30 & 107.71 \\
+\ours & 19.85 & \textbf{92.40} & \textbf{83.70} & \textbf{107.90} \\ \midrule
\multicolumn{5}{c}{\galaxys} \\ \midrule
(no graph) & 18.88 & 90.70 & 80.70 & 104.58 \\
+\bcilp & 18.88 & 90.20 & 80.40 & 104.18 \\
+\ccilp & 18.70 & 89.80 & 80.50 & 103.85 \\
+$\text{\ours}^\dagger$ &  19.04 & \textbf{91.10} & 81.40 & 105.29 \\
+\ours &  \textbf{19.12} & \textbf{91.10} & \textbf{82.30} & \textbf{105.82} \\ \bottomrule
\end{tabular}
\caption{Results from the response generation experiment with Galaxy~\cite{he2022galaxy}, HDNO~\cite{wang2020modelling} and HDSA~\cite{chen2019semantically} as base models. The \score metric is computed by \score=\bleu+(\inform+\success)/2. $\text{\ours}^\dagger$ stands for the ablation model that excludes \scond graph from our \ours. As we can see, our graphs consistently improves all metrics except \bleu for all models, and we improve the official scores by a significant amount for each model.
}
\label{tab:responseresults}
\end{table}

\subsection{Task 2: End-to-end Response Generation}
\paragraph{Base Models.}
We use the three SOTA end-to-end dialogue models finetuned on \woz: \galaxy~\cite{he2022galaxy}, \hdno~\cite{wang2020modelling}, and \hdsa~\cite{chen2019semantically} as base models.
`   Note that for \galaxy, since we were unable to reproduce the official prediction using the official repository, we report the result with both official prediction (\galaxy) and the greedy (\ie, beam search with beam width=1) prediction we obtained by running the official repository (\galaxys). 
\paragraph{Evaluation protocol.}
From each model, we first sample five system response utterances: one from official prediction\footnote{https://github.com/Tomiinek/MultiWOZ\_Evaluation/tree/
master/predictions}~\cite{nekvinda2021shades} and four from the model downloaded from the official implementation (see appendix~\ref{app:alternates} for details). 
The graph conditioning process follows \Cref{subsec:gce2e}. 
As an ablation, we also evaluated our method without conditioning on \scond graphs.
\paragraph{Metric.}
We follow the standard evaluation metric using the official code~\citep{nekvinda2021shades}, which computes three metrics on the \woz test set: \bleu (average \bleu~\citep{papineni-etal-2002-bleu} scores between generated and ground truth response), \inform (percentage of dialogs where the system presents an appropriate entity) and \success (percentage of dialogs where the task goals are achieved). The combined score is computed as \score=$\text{\bleu} + (\text{\inform} + \text{\success})/2$. See Appendix~\ref{app:multiwozeval} for details on computing these metrics. We report all four metrics of the compared methods. 

\paragraph{Result.}
We show the results in Table~\ref{tab:responseresults}. 
We found that \ours can consistently improve \inform and \success metrics for all the base end-to-end dialog models.
\bleu score fluctuates because it depends a lot on the exact wording of each response, which our graphs have no control over. 
The combined score consistently improves by 0.72, 0.49, 0.58, and 1.24 for \hdsa, \hdno, \galaxy, and \galaxys, respectively by conditioning with our \ours. Note that these score improvements are actually quite significant, as the difference between top and second top SOTA methods (\galaxy ~and HDNO) is only 0.74.

Next, we compare different graph generation methods: \bcilp, \ccilp, and \ours.
Overall, \ours consistently outperforms other baselines, \ccilp and \bcilp.
In fact, the \ccilp and \bcilp often underperforms the greedy prediction without graph conditioning (\ie, (no graph).
Since the sampled predictions are in general much worse than the greedy prediction, unless the graph-based ranking is highly accurate, it often samples the prediction that is worse than greedy prediction. 
On the contrary, \ours~is often able to accurately pick out the non-greedy better alternative to outperform the greedy predictions.
We show one such example in Table~\ref{tab:rgexample}.
We also found that excluding \scond graphs (\ie, $\text{\ours}^\dagger$) significantly reduced the improvement in all base models except \hdno. This shows that, while using only \ccond-\sncond graphs also consistently improves performance of all base models, including \scond graphs can generally get even better results.

\begin{table*}[t]
\centering
\scalebox{0.98}{
\begin{tabular}{lll}
\toprule
\multicolumn{1}{l}{\tb{History}} & \multicolumn{2}{l}{
\begin{tabular}[c]{@{}l@{}} ... (A chat between USER and SYSTEM)\\ \textit{USER: No, I’m not ready to book it yet. Thank you!}\end{tabular}
}\\
\hline
\multicolumn{1}{c}{} & \multicolumn{1}{c}{\tb{Utterance}} & \tb{Actions} \\ 
\hline
\multicolumn{1}{l}{\begin{tabular}[c]{@{}l@{}}\tb{GALAXY}\end{tabular}} & \multicolumn{1}{l}{\begin{tabular}[c]{@{}l@{}}\textit{SYSTEM: The address is {[}value\_address{]}, postcode}\\ \textit{{[}value\_postcode{]}, the phone number is }\\ \textit{{[}value\_phone{]}}\end{tabular}} & {\color[HTML]{333333} \begin{tabular}[c]{@{}l@{}}SYSTEM inform addr\\ SYSTEM inform post\\ \textcolor{red}{SYSTEM inform phone}\end{tabular}} \\ 
\hline
\multicolumn{1}{l}{\begin{tabular}[c]{@{}l@{}}
\tb{GALAXY}\\\tb{ + \ours}
\end{tabular}} & \multicolumn{1}{l}{\textit{SYSTEM: Is there anything else I can help you with?}} & \textcolor{green}{SYSTEM reqmore} \\
\hline
\multicolumn{1}{l}{\tb{Ground truth}} & \multicolumn{1}{l}{\begin{tabular}[c]{@{}l@{}}\textit{SYSTEM: Just let us know when you are ready and we}\\ \textit{will be happy to assist you. Have a great day}\end{tabular}} & \begin{tabular}[c]{@{}l@{}}SYSTEM reqmore\\ SYSTEM offer booking\end{tabular} \\ 
\bottomrule
\end{tabular}
}
\caption{Qualitative comparison of generated responses of \tb{GALAXY} and \tb{GALAXY+\ours} on the fourth turn of dialogue \texttt{SNG0772} in MultiWOZ.
\tb{GALAXY}'s greedy-sampled response informs information such as phone number and postcode, which is out of context.
Instead, our \ours chooses an alternate response that does not violate \ccond (\ie, {\color{red}SYSTEM inform phone}) and complies with \scond (\ie, {\color{green}SYSTEM reqmore}), and it turns out to be better and closer to the ground truth.
}
\label{tab:rgexample}

\end{table*}

\section{Related Work}\label{sec:related}

\paragraph{Task-oriented Dialog Systems.}
There are two main classes of task-oriented dialog systems: \emph{pipeline} systems and \emph{end-to-end} systems.
In \emph{pipeline} approaches, the dialog system is segmented into various modules such as natural language understanding (NLU), dialog state tracking (DST), dialog policy and natural language generation (NLG)~\citep{zhang2020recent}.
The NLU module first converts user language input into standardized dialog acts, slots and values.
The DST module keeps track of the current dialog state in terms of the dialog acts, slots and values.
Based on the current dialog state, the dialog policy module predicts the next action.
Previous works have viewed the dialog policy task as a Markov Decision Process (MDP)~\cite{kwan2023survey}. 
Common approaches include reinforcement learning methods such as Q-learning, policy gradient~\cite{lipton2017bbqnetworks,zhou2017endtoend,gordon-hall-etal-2020-learning} with experience-replay~\cite{malviya2023experience}, or model-based planning~\cite{peng2018deep}.
Lastly, given the current dialog states and the predicted dialog acts, the NLG module generates a response in natural language.
We integrate the \ours graph into the dialogue policy module to improve its action prediction performance by learning what a model can, should and should not predict.
%

On the other hand, \emph{end-to-end systems} integrate all functionalities into one module.
Many end-to-end models employ a singular language model to execute all four steps \citep{he2022space}.
Alternatively, other methods bypass certain steps, generating the final response directly~\citep{he2022galaxy, wang2022Task}.
Pipeline systems are more interpretable and modular, allowing independent updates for enhanced control. 
Ours work uses the TOD-Flow graph to enhance end-to-end model responses by selecting the best one based on alignment with learned \ccond, \scond and \sncond conditions.

\paragraph{Graph-based Dialog Systems.}
There has also been previous works attempting to integrate graphs into task-oriented dialog systems. 
Most of them focused on using graphs to represent or select information from a knowledge base~\cite{yang-etal-2020-graphdialog,yang2021gks}. 
TGDM~\cite{Choi2016UsingAD} attempted to create a dialog policy through manually constucted graphs, while a followup work~\cite{Kwon2018TaskGB} proposed a rule-based system for automatically inferring these graphs given a working dialog policy. 
\citet{raghu2021end} tackles the task-oriented dialogue problem where the system must ground dialog utterances to the manually-defined flowcharts describing the procedure and adapt to unseen ones during testing. 
\citet{laradji2023workflow} aims to discover the workflow, a sequence of dialog acts with their respective slot values, from unseen conversation. 
Our \ours graph is constructed from labeled dialogue data without any additional human supervision. 
And our graph models the relationship between dialog acts and slots; \eg, what dialog act or updates in slot values can happen or not.


\section{Conclusion}
This work introduced a novel framework for improving the efficiency and predictive accuracy of task-oriented dialogues models. By leveraging the concept of subtask graph and generalizing it to a \emph{TOD-flow graph}, we accurately inferred the latent task structure within a dialogue. As showcased through extensive experimentation with two public TOD datasets, the proposed technique has been proven to effectively generate accurate and human-interpretable graphs. Importantly, we have integrated these inferred graphs with a range of dialogue models, without necessitating retraining, resulting in a substantial enhancement in performance in both dialog act classification and end-to-end response generation.

\newpage

\section*{Limitations}

Although our method can be directly used when there are multiple domains involved in a single dialog (by treating combination of domains as a single separate domain and creating graphs using dialogs that has the same combination of domains, similar to what we did for MultiWOZ), this approach is limited in that (1) we need to infer a graph for every combination of domains that is present (such as the multi-domain dialogs in SGD, where there are 24 different single-domains alone), and (2) we cannot easily generalize to unseen domain combinations (even if we have graphs for each individual domain). In the future, we would like to explore ways to directly combine graphs for individual domains into multi-domain graphs and thus address the two limitations above.

We also relied on action annotations from the datasets to infer graphs, which limits the applicability of our approach.
It would be interesting to extend our approach to unannotated raw dialogues. 

\section*{Acknowledgments}
This work was supported in part by grants from LG AI Research.



\bibliography{anthology,custom}
\bibliographystyle{acl_natbib}
\appendix

\section{Experiment Details}
\label{app:expdetail}
\subsection{Next Action Prediction}

\begin{figure*}[t]
\centering
\includegraphics[width=\textwidth]{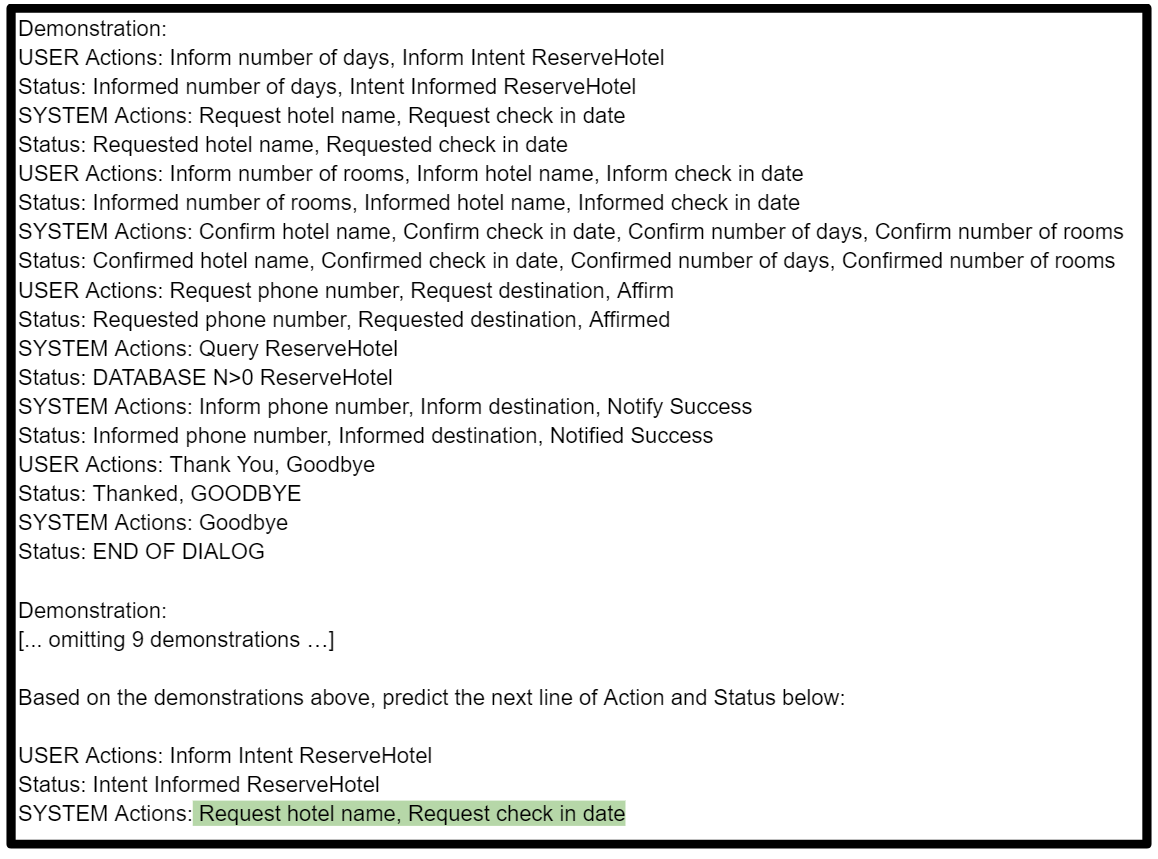}
\includegraphics[width=\textwidth]{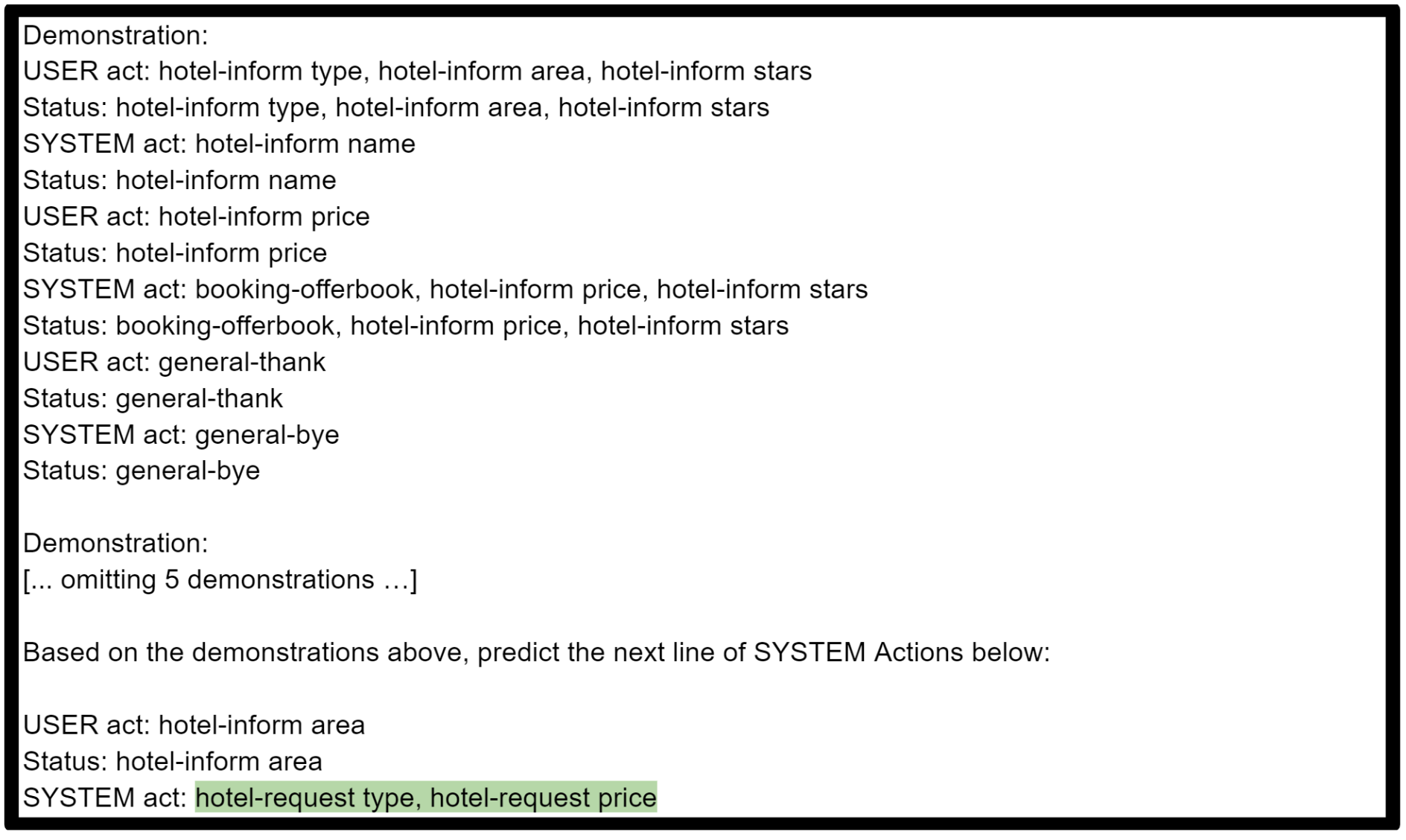}
\caption{Example of using GPT-turbo-3.5 to do next system action prediction from the Hotels\_1 domain of SGD (top) and from the Hotel+Train domain of MultiWOZ (bottom). We first provide a few demonstration trajectories in the prompts, then we ask the large language model to complete the prompt. The text highlighted in green is the GPT completion, while everything before that is our input prompt. We give the exact same prompt to the other LLM (FLAN-T5-xxl).}
\label{fig:promptexample}
\end{figure*}

\subsubsection{Dataset Preprocessing details}
\label{app:dp}

In the next action prediction experiment, we are using 2 datasets: MultiWOZ~\cite{budzianowski2020multiwoz} and SGD~\cite{rastogi2020towards}. We first split each dataset into domains and train/test splits. For MultiWOZ, the dataset has an official splitting of train/val/test splits, so we follow the same splits; while MultiWOZ contains dialogs from 7 domains, 2 of them (police, hospital) have no test dialogs, thus we split MultiWOZ dialogs into 14 domains, including 5 single-domains (i.e. dialogs that only involves one of the 5 domains) and 9 multi-domains (i.e. dialogs that involves multiple domains, such as Hotel+Train). For SGD, since the official train/val/test splits often involves test schemas that does not exist in the train set, we decided to create our own train/test splits from the official training set. There are 24 different schemas that have single-schema dialogs in the official training set, so we treat each of these schemas as a separate domain and randomly split dialogs within each domain into train/test splits at a 9:1 ratio.

We then turn each dialog into a trajectory (as defined in section~\ref{sec:problem}). Below is how we define the dialog actions within each domain of each dataset:

For SGD, since there are already very comprehensive dialog acts and slot annotations, we directly use the acts defined in the dataset (with a few re-naming) plus the slot annotations to build our set of all possible actions. In addition, since SGD provides explicit annotations about system's database queries, whenever the system queries database, we add an additional turn in our trajectory with one action "SYSTEM query <Intent>" and the status update would be either query success or query failure. 

For MultiWOZ, we mostly also directly use the acts and slots in the dataset annotation to build our set of all possible actions, but we did some re-naming and re-organization to remove some redundant combinations of acts and slots (for example, "SYSTEM Booking-Inform <slot>" was changed into "SYSTEM OfferBook + SYSTEM inform <slot>"). Since there are no explicit annotation about database querying, we simply assume that the system looks up information before each "book/nobook" operation and add a corresponding status update to the utterance before these actions. We do not explicitly add "query" actions or additional turns to the trajectories. 

Then, within each domain, we will use the trajectories of the train split dialogs to obtain graphs and also act as demonstrations for LLMs, and use the graphs to improve next action predictions on the test dialogs.

\subsubsection{Large Language Model Prompting Details}
\label{app:prompt}

In this experiment, we used GPT-turbo-3.5 as well as FLAN-T5~\cite{chung2022scaling} as baseline next action predictors. We prompt the two LLMs using the exact same prompting method. For each domain, we first randomly select 10 dialogs from the training split, and then use their trajectories (i.e. actions and statuses of each turn) as demonstrations. Then for every system utterance in the test-split dialogs in the domain, we include as many of the demonstration trajectories as possible without exceeding the max token limit of the LLMs, and then we include the partial trajectory of the test dialog up to the turn where the next system actions needs to be predicted. Lastly, we ask the LLM to predict the next system actions, and we programmatically parse the results into individual action items. See Figure~\ref{fig:promptexample} for example prompt for SGD and MultiWOZ dataset.

When obtaining the baseline result for each model, we use the top prediction by probability by setting temperature to zero (thus the language model's generation is deterministic). When we need to multiple predictions for graph filtering, for FLAM-T5-xxl we simply do a beam-search of size 10; for GPT-turbo-3.5 we set temperature to 1 and sample 10 times.

\subsection{Response Generation}

\subsubsection{Getting Alternates from each model}
\label{app:alternates}

For Galaxy~\cite{he2022galaxy}, we used the top choice of beam-size-1 as well as the top 3 choices of beam-size-5 as alternates. The overall ranking of the 5 choices from highest to lowest are baseline, beam-size-1, beam-size-5 top choice, beam-size-5 second choice, beam-size-5 third choice.

For HDNO~\cite{wang2020modelling}, since the official prediction baseline is the top choice of beam-size-5, we use the two choices from beam-size-2 and the second and third choice from beam-size-5 as alternates. The overall ranking of the 5 choices from highest to lowest are baseline (i.e. beam-size-5 top choice), beam-size-2 top choice, beam-size-5 second choice, beam-size-2 second choice, beam-size-5 third choice.

For HDSA~\cite{chen2019semantically} the process is slightly different. HDSA model consists of 2 parts: the first part (predictor) predicts the actions the system will perform (although in a very different format than what we do in our next action prediction experiment, so not directly comparable), and the second part (generator) uses the output of the predictor to generate the response. If we fix the predictor output and do beam-search on the generator only, the actions within the generated response will almost always be identical, which renders our method useless. Therefore, we created our alternates by tweaking a hyperparameter in the predictor a little bit. The output of the predictor is a binary vector, and the post-sigmoid logits of the predictor is converted to the binary vector by a threshold. The HDSA official code repository has 0.4 as the default threshold, and we changed the threshold around that value and used the different generated vectors as inputs to the generator to obtain our alternates. The overall ranking of the 5 choices from highest to lowest are baseline, threshold-0.4, threshold-0.375, threshold-0.35, threshold-0.325.

\subsubsection{Evaluation Details}
\label{app:multiwozeval}

We use the official MultiWOZ\_Evaluation repository to evaluate the BLEU/INFORM/SUCCESS metrics. Since our policy-learning setting assumes that we have access to the ground truth dialog state before the utterance, we use the ground truth dialog state and active domains in the evaluation scripts (by removing dialog state / active domain predictions and only including the response in the prediction file). This is necessary because we found that active domain predictions affect the INFORM/SUCCESS metrics, and incorrect active domain can increase/decrease INFORM/SUCCESS randomly. Therefore, to ensure fairness and consistency, we always use the ground truth active domain during evaluation.

\begin{figure*}[h!]
\centering
\includegraphics[width=\textwidth]{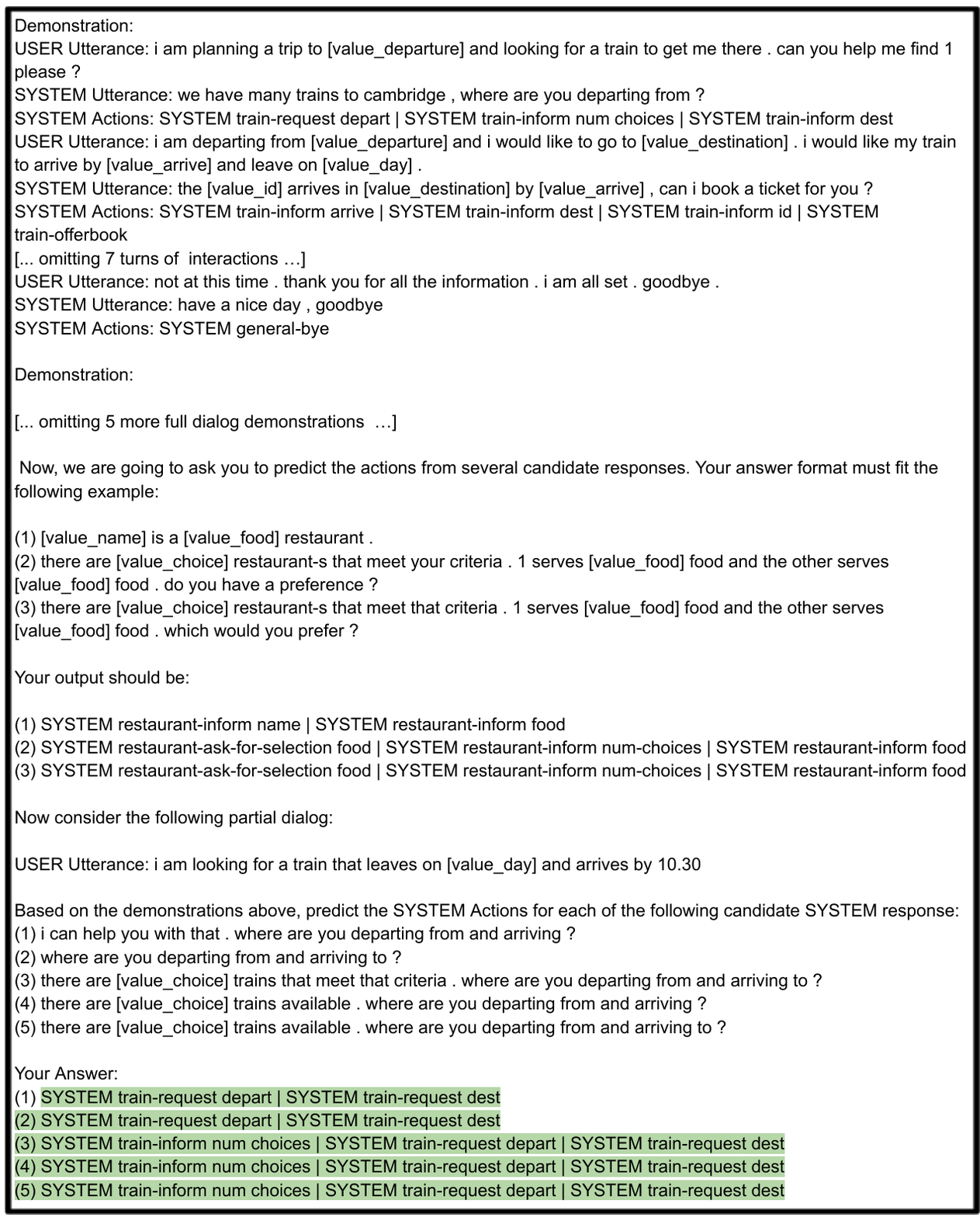}
\caption{Example of using GPT-turbo-3.5 to obtain actions from candidate responses from Galaxy~\cite{he2022galaxy}. The parts highlighted in green are completion from GPT, and everything before that is our prompt.}
\label{fig:nlupromptexample}
\end{figure*}
\subsubsection{Using GPT as NLU unit}
\label{app:nlu}

We prompt GPT-turbo-3.5 to convert the candidate responses into action sets. For responses for dialogs in each domain, We first provide randomly selected dialogs from the training split of the same domain together with their ground truth system actions as demonstrations. Then, we specify the desired output format and provide an example of the output format. Lastly, we provide the dialog history of the current candidate responses, and ask GPT to give us actions to all candidate responses (i.e. the baseline + 4 sampled from the models). We start with 6 demonstrations, and we reduce the number of demonstrations by one iteratively if the total number of tokens exceeds the maximum token limit of the model (4097). We show one example prompt for responses from Galaxy~\cite{he2022galaxy} together with the GPT completion in Figure~\ref{fig:nlupromptexample}.

We evaluated the quality of this NLU process by using this process to predict actions of the ground truth responses and compare the predicted actions to the ground truth actions on a subset of the test dialogs. We found that on average our NLU's predicted actions achieves an average F-1 score of 77.6\%, which is okay but far from perfect, and the imperfectness of our NLU brings additional challenge to our task.

\section{Qualitative comparison between inferred and human-drawn graphs}
\label{sec:gtgraph}

\begin{figure*}[h!]
\centering
\includegraphics[width=\textwidth]{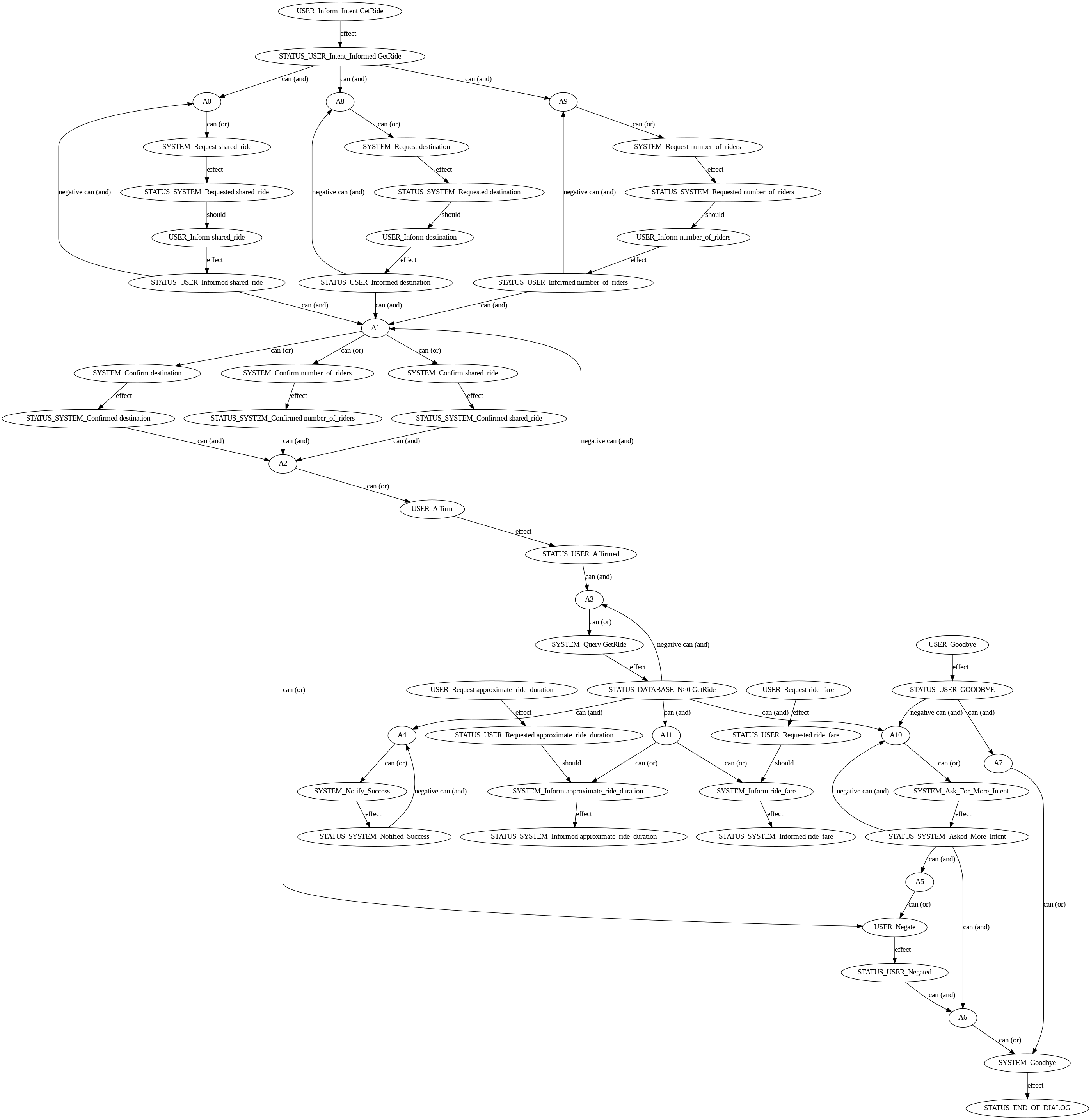}
\caption{Human-drawn ground truth graph for RideSharing\_1 domain in SGD.}
\label{fig:gt1}
\end{figure*}

\begin{figure*}[h!]
\centering
\includegraphics[width=\textwidth]{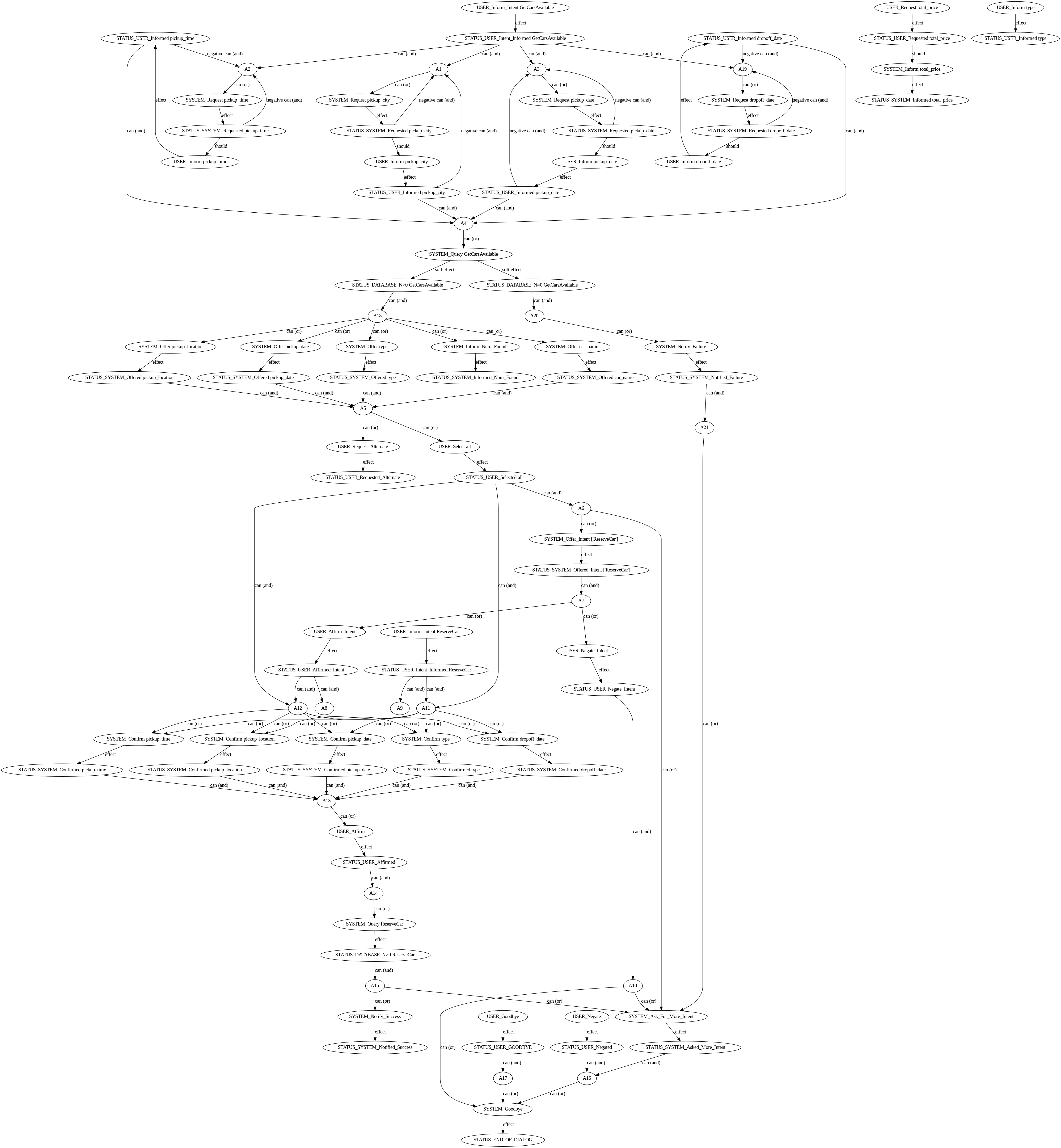}
\caption{Human-drawn ground truth graph for RentalCars\_1 domain in SGD.}
\label{fig:gt2}
\end{figure*}

\begin{figure*}[h!]
\centering
\includegraphics[width=\textwidth]{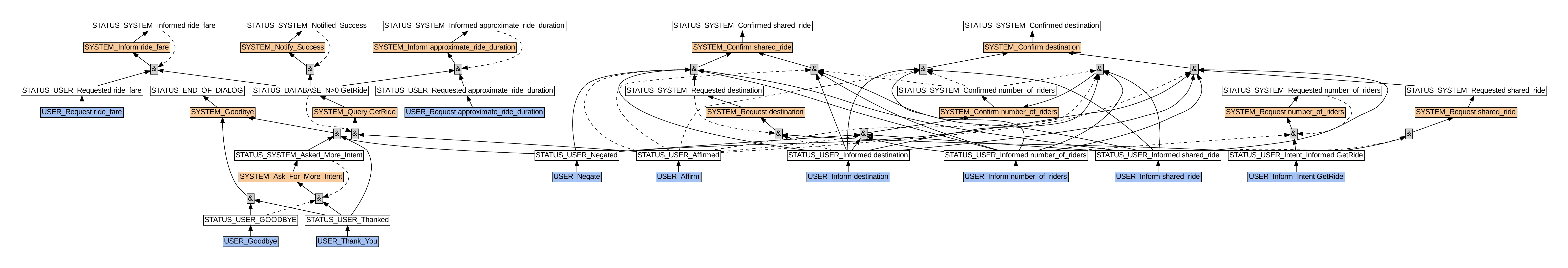}
\caption{Inferred \ours graph of system dialog acts for RideSharing\_1 domain in SGD.}
\label{fig:infer1}
\end{figure*}

\begin{figure*}[h!]
\centering
\includegraphics[width=\textwidth]{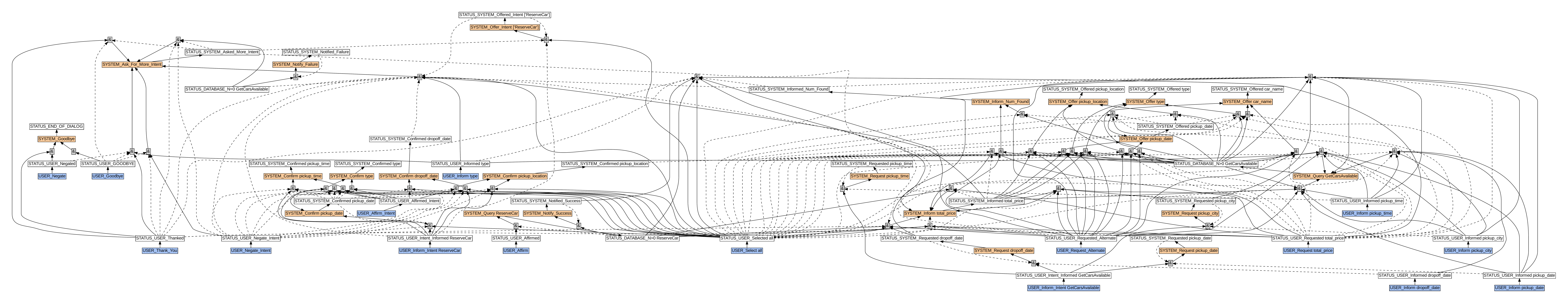}
\caption{Inferred \ours graph of system dialog acts for RentalCars\_1 domain in SGD.}
\label{fig:infer2}
\end{figure*}

In order to perform qualitative assessment of the quality of our graphs, we manually drew graphs for RideSharing\_1 and RentalCars\_1 domains in SGD. 
Then, we compare the manually drawn graph with the inferred graphs in Figure~\ref{fig:gt1} to Figure~\ref{fig:infer2}.
Note that for readability, we only visualized the \ccond and \sncond (\ie, no \scond) of system dialog acts (\ie, no user dialog acts) in the figures.
In general, the inferred graphs looks more complicated (\ie, more edges) than the human-drawn graphs.
This is mainly due to small annotation errors in the data.
For example, \texttt{NOT}(\texttt{USER\_Negated})\&\texttt{USER\_Affirmed} is equivalent to \texttt{USER\_Affirmed} in Boolean algebra, and human tends to choose latter.
But when there are few annotation errors in the trajectory data for either \texttt{USER\_Negated} or \texttt{USER\_Affirmed}, the inferred graph will try to incorporate both (\ie, consider both as the precondition) to make a best guess about whether the user actually negated or affirmed.
Thus, most of these added preconditions are in fact redundant, and they do not affect the prediction performance of downstream dialog policy or end-to-end models.

\end{document}